\documentclass[10pt,twocolumn,letterpaper]{article}

\usepackage{cvpr}
\usepackage{times}
\usepackage{epsfig}
\usepackage{graphicx}
\usepackage{amsmath}
\usepackage{amssymb}
\usepackage{subcaption}
\usepackage[utf8]{inputenc}
\usepackage{CJKutf8}
\usepackage{booktabs}
\usepackage{longtable}
\usepackage{multirow}
\usepackage{array}
\usepackage{adjustbox}
\usepackage{placeins} 
\usepackage{float}
\usepackage{xcolor}

\usepackage[pagebackref=true,breaklinks=true,letterpaper=true,colorlinks,bookmarks=false]{hyperref}

\cvprfinalcopy 

\def\cvprPaperID{****} 

\ifcvprfinal\fi
\begin{document}

\title{Efficient Adaptation of Multilingual Models for Japanese ASR}

\author{
Mark Bajo \quad Haruka Fukukawa \quad Ryuji Morita \quad Yuma Ogasawara\\
Georgia Institute of Technology\\
\tt\small mbajo3@gatech.edu, hfukukawa3@gatech.edu, ryuji\_morita@gatech.edu, yogasawara3@gatech.edu\\
}

\maketitle

\begin{abstract}
This study explores fine-tuning multilingual ASR (Automatic Speech Recognition) models, specifically OpenAI’s Whisper-Tiny, to improve performance in Japanese. While multilingual models like Whisper offer versatility, they often lack precision in specific languages. Conversely, monolingual models like ReazonSpeech excel in language-specific tasks but are less adaptable. Using Japanese-specific datasets and Low-Rank Adaptation (LoRA) along with end-to-end(E2E) training, we fine-tuned Whisper-Tiny to bridge this gap. Our results show that fine-tuning reduced Whisper-Tiny’s Character Error Rate (CER) from 32.7 to 20.8 with LoRA and to 14.7 with end-to-end fine-tuning, surpassing Whisper-Base’s CER of 20.2. However, challenges with domain-specific terms remain, highlighting the need for specialized datasets. These findings demonstrate that fine-tuning multilingual models can achieve strong language-specific performance while retaining their flexibility. This approach provides a scalable solution for improving ASR in resource-constrained environments and languages with complex writing systems like Japanese.
\end{abstract}

\section{Introduction/Background/Motivation}

Our project aims to improve Japanese ASR by fine-tuning a multilingual model, originally trained mainly on English, using Japanese texts to enhance its performance. In the Japanese ASR domain, models like ReazonSpeech, which are specifically trained on Japanese corpora\cite{huggingfaceReazonspeechDatasets}, excel in capturing language-specific nuances and demonstrate strong baseline performance. However, these models lack versatility across different languages.

Conversely, OpenAI's Whisper model is trained on multiple languages\cite{radford2023robust}, offering broad applicability but often lacking deep specialization in any single language\cite{githubWhispermodelcardmdMain}. This presents a significant trade-off in the ASR landscape: while multilingual models provide versatility across languages, they may not match the precision of monolingual models for specific languages like Japanese. This is primarily because multilingual models have reduced per-language capacity compared to monolingual models of the same size. However, fine-tuning multilingual models like Whisper with additional language-specific data, such as Japanese, can significantly enhance their performance by increasing accuracy and adapting to linguistic nuances. This is supported by Whisper's demonstrated ability to improve performance on multilingual benchmarks like Multilingual LibriSpeech (MLS), where it outperformed other models such as XLS-R and mSLAM in a zero-shot setting, indicating that additional language-specific fine-tuning could further enhance its capabilities for specific languages\cite{radford2023robust}.

Our approach involves using LoRA \cite{huLoRALowRankAdaptation2021} alongside E2E fine-tuning of the Whisper model with a Japanese-specific dataset. LoRA freezes the pre-trained model weights and injects trainable rank decomposition matrices into each layer, which signficantly reduces the number of trainable parameters, leading to efficient training with smaller datasets, which is aligned with our task \cite{huLoRALowRankAdaptation2021}. The goal is to evaluate the extent of improvement achieved after fine-tuning compared to the performance of the same Whisper model prior to fine-tuning. For benchmarking purposes, we use the monolingual model ReazonSpeech\cite{githubReazonSpeech}, specifically one with a similar parameter size as the Whisper tiny or small models\cite{githubOpenaiWhisper}, to provide a point of comparison. However, our purpose is not to surpass ReazonSpeech but rather to explore how much improvement can be achieved compared to the baseline Whisper models.

The success of our project could have a transformative impact on ASR development, particularly for languages with complex writing systems like Japanese. Improved ASR models could provide tangible benefits across various sectors. For instance, individuals with disabilities who rely on ASR for communication, customer service systems requiring precise transcription and translation, and educational tools designed to support language learning could all benefit from enhanced speech recognition solutions. By bridging the gap between monolingual and multilingual performance, our project also aims to contribute valuable insights to the broader ASR community, influencing future research and development strategies.

\subsection{Dataset}
We used the Japanese datasets from Google Fleurs (GF)\cite{conneauFLEURSFewshotLearning2022}, Common Voice (CV)\cite{ardilaCommonVoiceMassivelyMultilingual2020}, JSUT\cite{sonobeJSUTCorpusFree2017}, and ReazonSpeech \cite{yinReazonSpeechFreeMassive} for training, and GF, CV and JSUT for testing, with a 80:10:10 train, validation, and test split. The Whisper models were previously evaluated in Japanese using CV and GF \cite{radfordRobustSpeechRecognition}, and the ReazonSpeech model was previously evaluated using the JSUT and CV \cite{holdingsReazonSpeechV21Setting2024}. Since the training sets used for the Whisper models were not disclosed, we leveraged the four datasets under the assumption that they were not already used. The datasets give us diversity in quality, vocal gender, and comprehensiveness of the language. A comparison of the datasets is summarized in Table~\ref{tab:comparisondatasets}.
\begin{table}[ht]
\centering
\resizebox{0.45\textwidth}{!}{
\begin{tabular}{|c|c|c|c|c|}
\hline
\textbf{Attribute} & \textbf{GF} & \textbf{CV} & \textbf{JSUT} & \textbf{ReazonSpeech} \\
\hline
\textbf{Background Noise} & \checkmark  & \checkmark  & - & \checkmark  \\
\hline
\textbf{Pauses} & \checkmark  & \checkmark  & - & -  \\
\hline
\textbf{Varying Volume} & \checkmark  & \checkmark  & - & -  \\
\hline
\textbf{Emotional Tones} & -    & \checkmark    & -   & \checkmark       \\
\hline
\textbf{Male Voices} & \checkmark    & \checkmark    & -   & \checkmark       \\
\hline
\textbf{Female Voices} & \checkmark  & \checkmark  & \checkmark  & \checkmark   \\
\hline
\textbf{Non-Native Speakers} & -    & \checkmark    & -      & -    \\
\hline
\end{tabular}
}
\caption{Comparison of Datasets}
\label{tab:comparisondatasets}
\end{table}

\textbf{Google Fleurs (GF):}
The dataset is described to have equal gender representation \cite{conneauFLEURSFewshotLearning2022}, but have more male voices than female voices. A limited number of speakers produced all samples.

\textbf{JSUT:}
JSUT contains a single female Japanese speaker, and is the only dataset recorded in a professional studio \cite{sonobeJSUTCorpusFree2017}. We will use the Basic5000 subcorpus of JSUT, which covers all readings of kanjis used in daily life. The advantage of this dataset is the clean audio and comprehensive construction of examples, but the limitation is the lack of variety in quality, gender, and tone. 

\textbf{Common Voice Japanese Corpus (CV):}
CV contains over 4000 voices of “native Japanese speakers”\cite{ardilaCommonVoiceMassivelyMultilingual2020}, however we found that a sizable portion of the voices, especially of the female voices, were not actually native Japanese speakers. The diversity and noise in the dataset may contribute to a more robust model that can perform ASR without pristine audio conditions and even for non-native speakers which may better reflect real-world usage. However, this non-uniformity in quality may make learning difficult. Ultimately, we decided that using this dataset in conjunction with the others, such as JSUT which provides clearly spoken native Japanese by a female speaker, would give balance to the overall dataset and contribute to the robustness of the model.

\textbf{ReazonSpeech:}
This is the only dataset that uses audio sampled from Japanese TV \cite{yinReazonSpeechFreeMassive} instead of recordings gathered for the purpose of a creating a dataset. As such, samples contain emotional tones, background music, fast speech, and both real and fantasy words. We did not include this in our test set because this was not used as a benchmark in other papers, and some examples contained words that only exist in anime.

\subsection{Models}

With the datasets defined, we now examine the architecture of the Whisper model, which serves as the backbone of our approach. Understanding its design and capabilities is crucial for contextualizing the fine-tuning process and evaluating its potential to improve the performance of Japanese ASR. Although Whisper is the main model in this study, we also used ReazonSpeech as a benchmark for Japanese ASR, providing a comparison to evaluate Whisper’s effectiveness in this context.

\textbf{Whisper:}
Whisper is a Transformer-based model designed for robust speech-to-text and language translation tasks\cite{IndroducingWhisper, MediumWhisper}. By training on 680,000 hours of diverse multilingual audio data, Whisper achieves significant performance improvements in scenarios involving noisy environments, accents, and specialized terminology. The model supports multiple tasks, including transcription, translation, and timestamping, without requiring separate architectures for each.

The Whisper model processes audio in 30-second segments, converting the input into log-Mel spectrograms to standardize its representation. These spectrograms are then passed through two convolutional layers with GELU activation, which help extract low-level features from the audio input. Following this, sinusoidal positional encodings are added to encode temporal information explicitly, ensuring the model can capture the sequence of audio events.
The processed features are then fed into Transformer Encoder Blocks, which consist of multiple layers of self-attention mechanisms and feed-forward networks(MLPs). These layers enable the encoder to extract hierarchical latent representations of the audio, capturing both local and global dependencies across the input sequence.
On the decoding side, the Transformer Decoder Blocks generate text tokens by attending to the encoder’s latent representations via cross-attention mechanisms. Each decoder block includes self-attention for understanding relationships between previously generated tokens, cross-attention for aligning to the encoder output, and MLPs for feature transformation. The decoder also incorporates learned positional encodings to ensure token predictions respect the sequential structure of the output text.
In Table~\ref{tab:comparison_whisper_models}, we present a concise comparison of the Whisper models, highlighting key architectural parameters.

\begin{table}[ht]
\centering
\resizebox{0.45\textwidth}{!}{
\begin{tabular}{|c|c|c|c|c|c|}
\hline
\textbf{Attribute} & \textbf{Tiny} & \textbf{Base} & \textbf{Small} & \textbf{Medium} & \textbf{Large} \\
\hline
\textbf{Number of Parameters} & 39M  & 74M  & 244M & 769M & 1550M \\
\hline
\textbf{Encoder Layers} & 4    & 6    & 12   & 24   & 32    \\
\hline
\textbf{Decoder Layers} & 4    & 6    & 12   & 24   & 32    \\
\hline
\textbf{Hidden Dimensions} & 384  & 512  & 768  & 1024 & 1280  \\
\hline
\textbf{Attention Heads} & 6    & 8    & 12   & 16   & 20    \\
\hline
\end{tabular}
}
\caption{Comparison of Whisper models}
\label{tab:comparison_whisper_models}
\end{table}

Smaller models like Whisper-Tiny are ideal for resource-constrained tasks, while larger models like Whisper-Large offer higher accuracy, making them ideal for demanding tasks that require substantial processing power. 

\textbf{ReazonSpeech:}
To further benchmark and compare Whisper’s performance, we introduce the ReazonSpeech models, which are specifically designed for Japanese ASR. ReazonSpeech models offer an alternative approach to Japanese speech recognition, offering a valuable comparison to Whisper’s Transformer-based architecture. By examining these models, we can better understand how the performance of our model in Japanese ASR compares to other established systems. The three ReazonSpeech models—ReazonSpeech-nemo\cite{ReazonSpeechNemo}, ReazonSpeech-k2\cite{ReazonSpeechK2}, and ReazonSpeech-espnet\cite{ReazonSpeechESPNet}—each utilize distinct techniques and optimizations, making them suitable for evaluating accuracy differences relative to our model.

\section{Approach}

\begin{figure*}[t]
    \centering
    \begin{subfigure}[b]{0.24\textwidth}
        \centering
        \includegraphics[width=\textwidth]{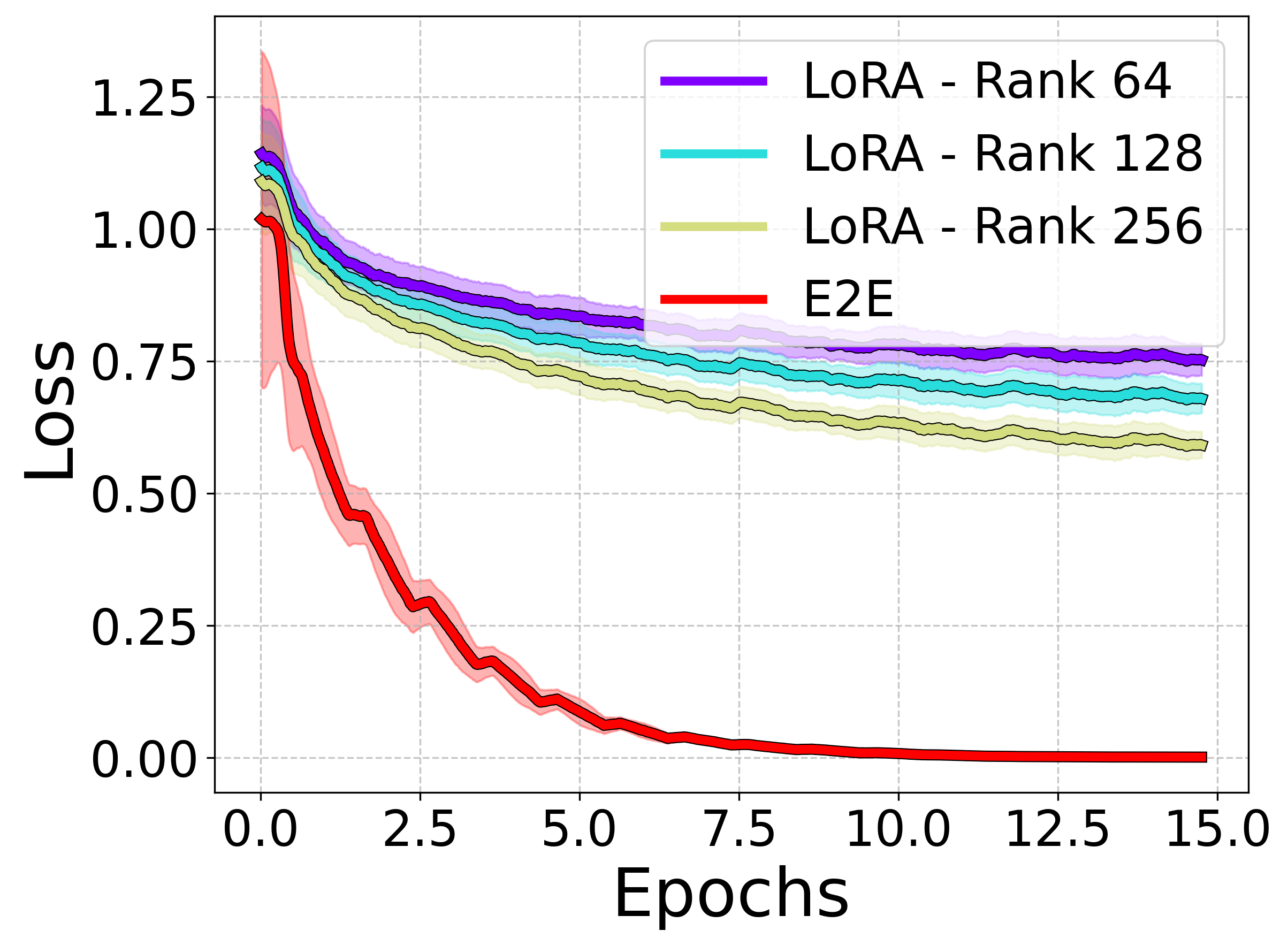}
        \caption{Training Loss Curve}
    \end{subfigure}
    \begin{subfigure}[b]{0.24\textwidth}
        \centering
        \includegraphics[width=\textwidth]{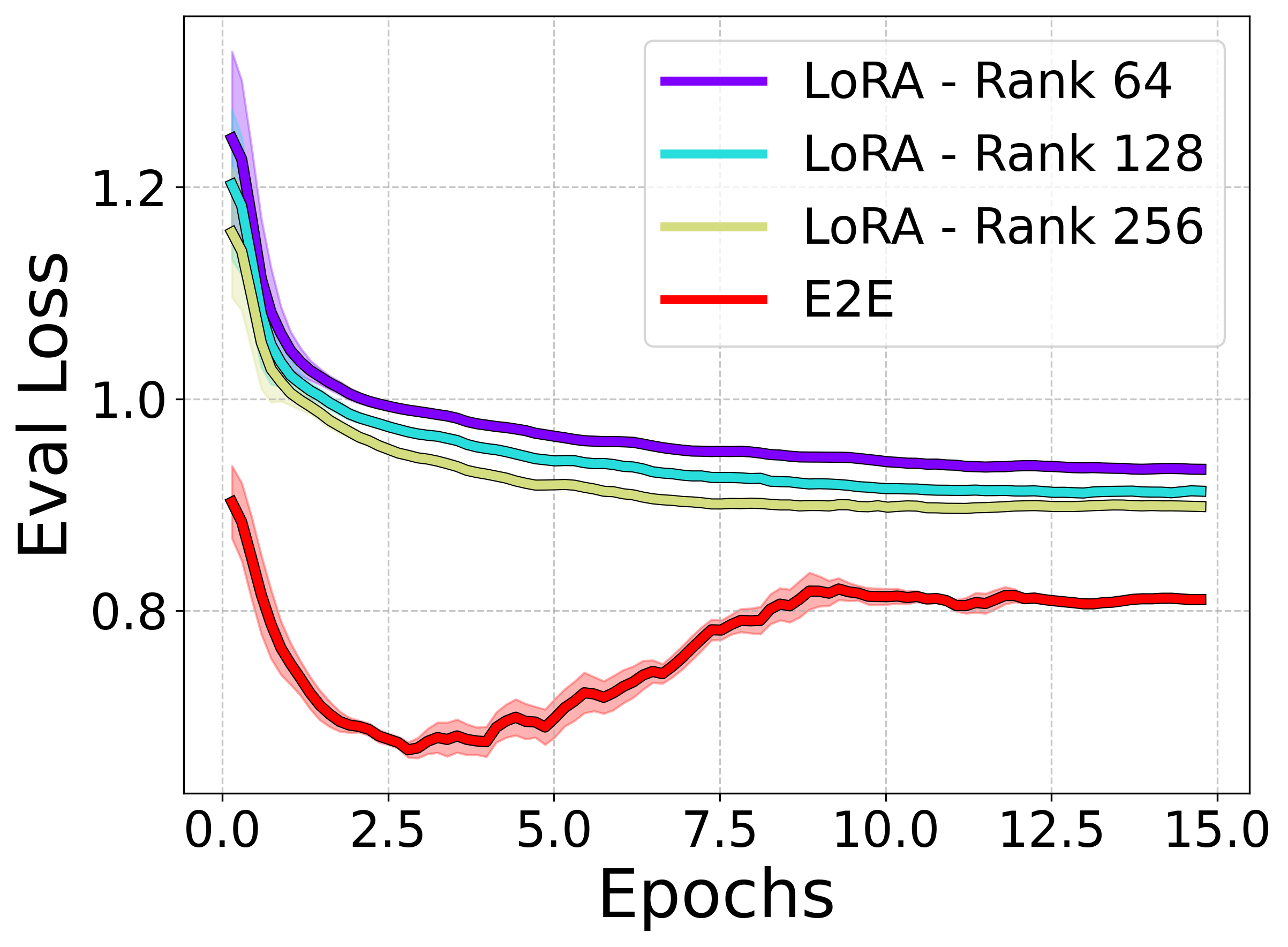}
        \caption{Eval Loss Curve}
    \end{subfigure}
    \begin{subfigure}[b]{0.24\textwidth}
        \centering
        \includegraphics[width=\textwidth]{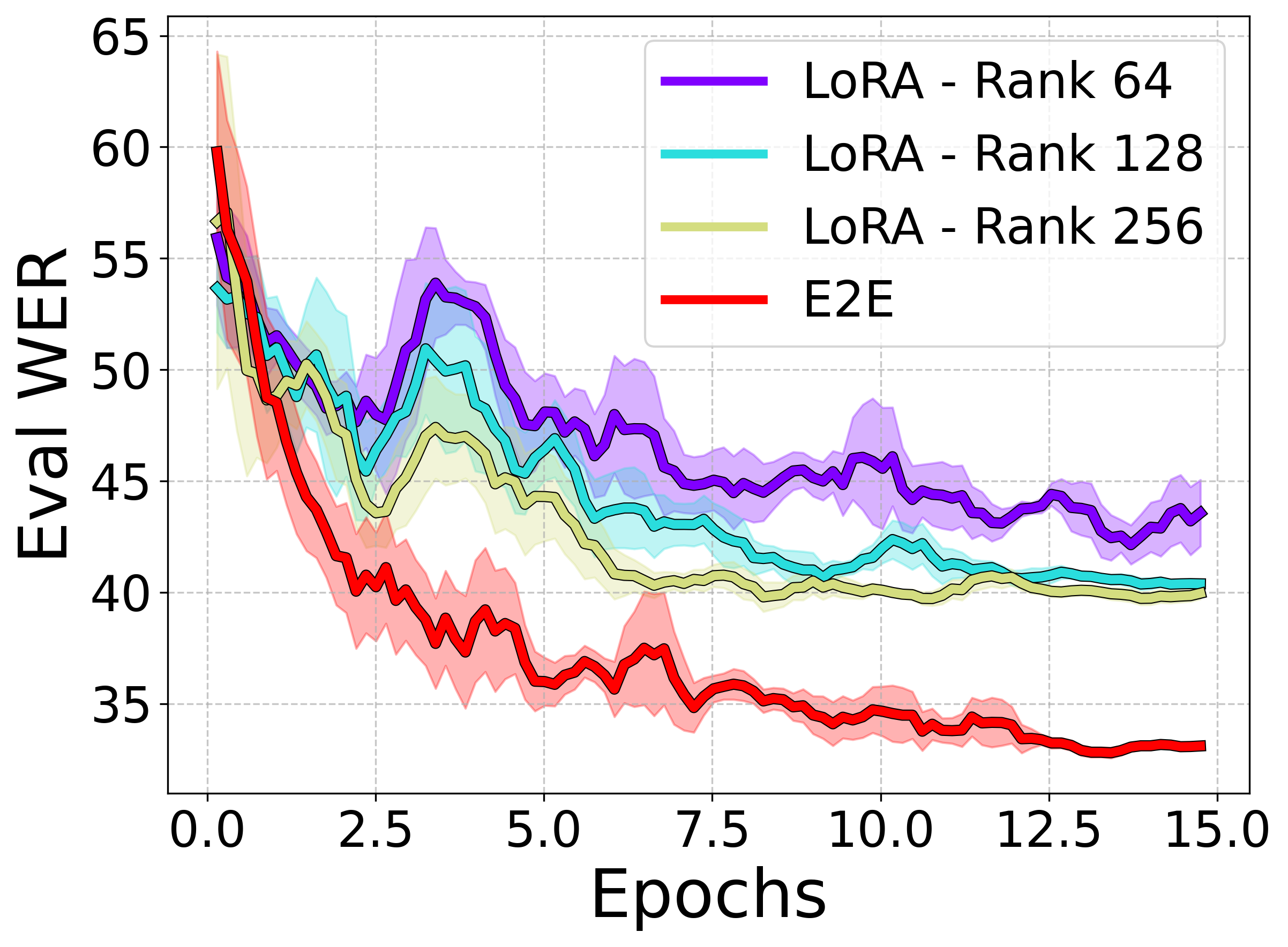}
        \caption{Eval WER Curve}
    \end{subfigure}
    \begin{subfigure}[b]{0.24\textwidth}
        \centering
        \includegraphics[width=\textwidth]{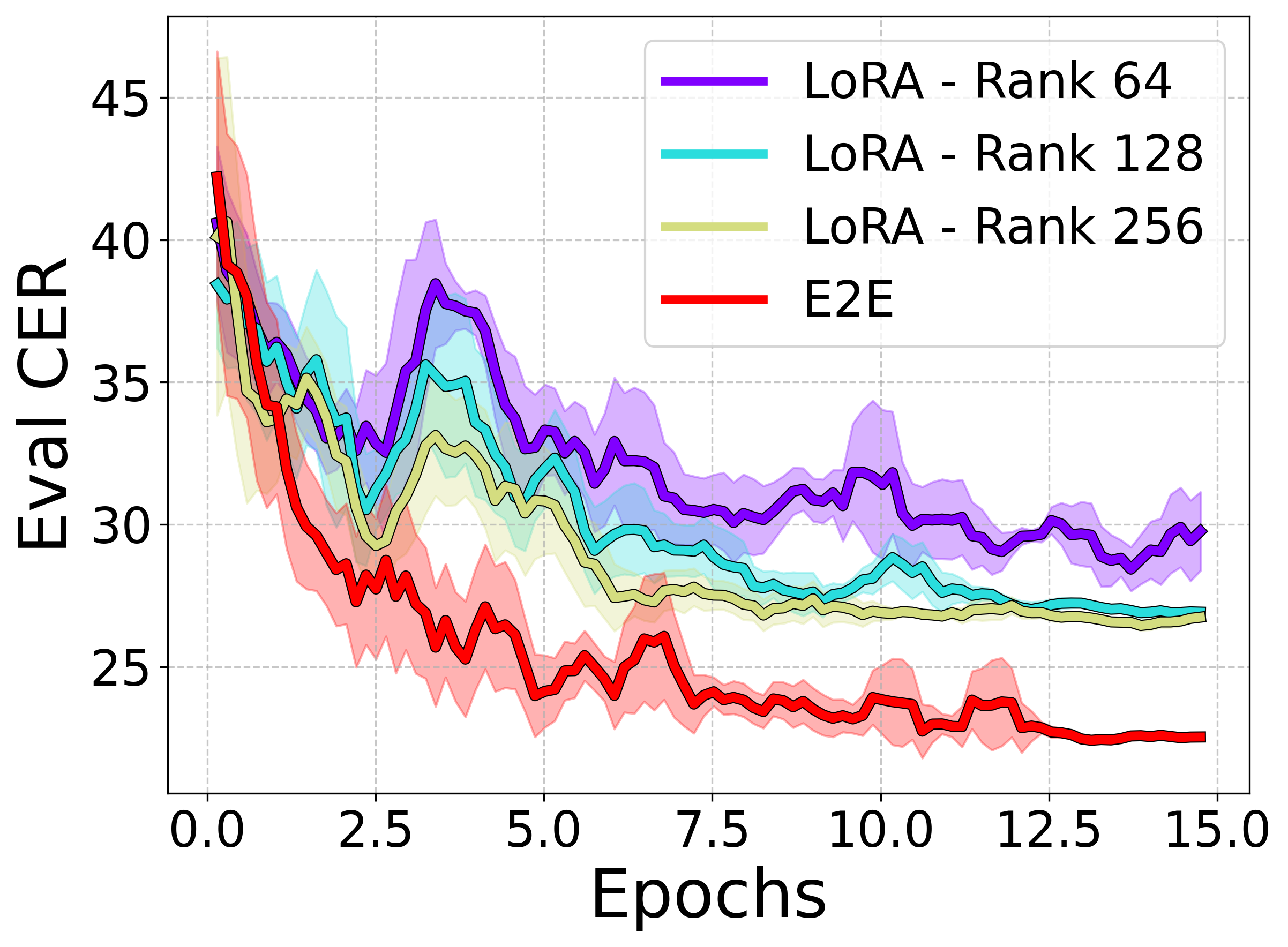}
        \caption{Eval CER Curve}
    \end{subfigure}
    \caption{Training and evaluation curves for different Rank values of LoRA. We also show the performance of the end-to-end fine-tuning approach. These curves are from fine-tuning the Tiny model.}
    \label{fig:lora_comparison}
\end{figure*}

This research aims to enhance the Whisper-Tiny model for Japanese ASR through fine-tuning approaches. We explored two primary methods: LoRA \cite{hu2021loralowrankadaptationlarge,xu2023parameterefficientfinetuningmethodspretrained} and end-to-end fine-tuning, focusing on Whisper-Tiny for its resource efficiency.

We anticipated three main challenges: managing memory constraints while fine-tuning in large datasets, preventing overfitting due to limited or imbalanced data, and handling the complexity of Japanese writing systems, where multiple valid representations(e.g., kanji and hiragana) can lead to inconsistencies in transcription. These considerations shaped our approach to data preparation and model optimization.

Our methodology combined four Japanese datasets (GF, CV, JSUT, and ReazonSpeech) into a comprehensive training set, implementing SpecAugment for data augmentation. Traditional audio augmentation techniques like time-stretching and pitch-shifting were also tested, but ultimately excluded due to minimal impact.

For LoRA, we modified key transformer layers with rank-adaptive matrices, experimenting with rank values to optimize performance while preserving most original parameters. In parallel, we conducted end-to-end fine-tuning, updating all parameters using our combined dataset. This involved carefully implementing gradient checkpointing and learning rate scheduling to address memory constraints and avoid overfitting.

Several significant challenges arose during our research. Memory constraints initially prevented training larger models, which we addressed through gradient checkpointing and batch size optimization. Overfitting posed another challenge, as the model exhibited strong training performance but poor generalization. This issue was successfully mitigated through the use of data augmentation and careful tuning of weight decay parameters. The CV dataset included invalid examples, requiring filtering to create an effective training set.

Initial attempts revealed important insights that shaped our final approach. We started with direct fine-tuning of Whisper-Tiny but encountered immediate overfitting issues, leading us to explore alternative strategies. Through iterations, we introduced LoRA to address memory constraints and added SpecAugment after observing limited generalization. Some experiments, such as traditional audio augmentation and larger batch sizes, showed minimal benefits or decreased performance, guiding us to focus our efforts on more effective approaches. Through these iterations and challenges, we developed a robust methodology that effectively improved Japanese ASR performance while maintaining practical resource requirements.

\section{Experiments and Results}

\begin{figure*}[t]
    \centering
    \begin{subfigure}[b]{0.24\textwidth}
        \centering
        \includegraphics[width=\textwidth]{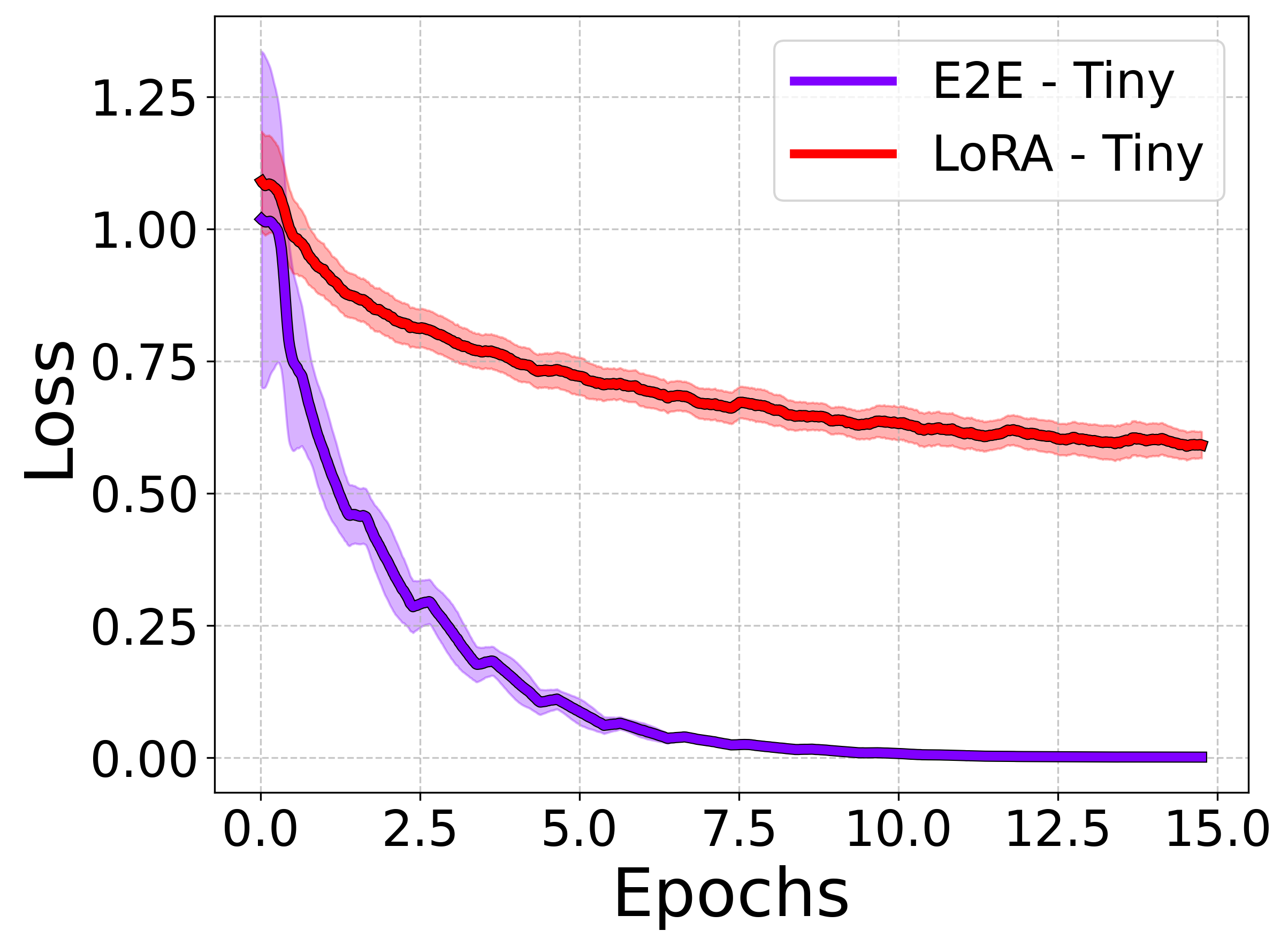}
        \caption{Tiny Train Loss}
    \end{subfigure}
    \begin{subfigure}[b]{0.24\textwidth}
        \centering
        \includegraphics[width=\textwidth]{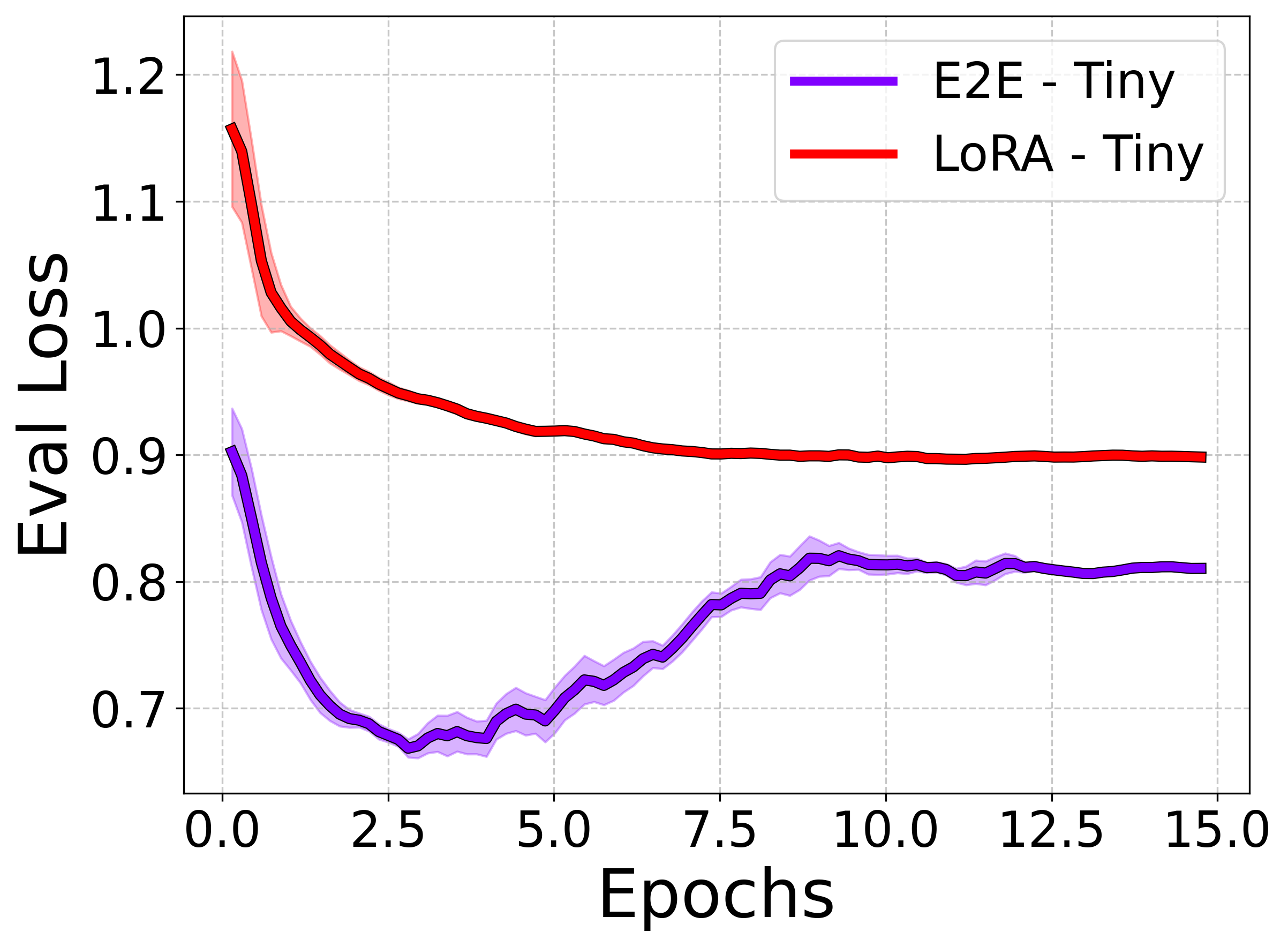}
        \caption{Tiny Eval Loss}
    \end{subfigure}
    \begin{subfigure}[b]{0.24\textwidth}
        \centering
        \includegraphics[width=\textwidth]{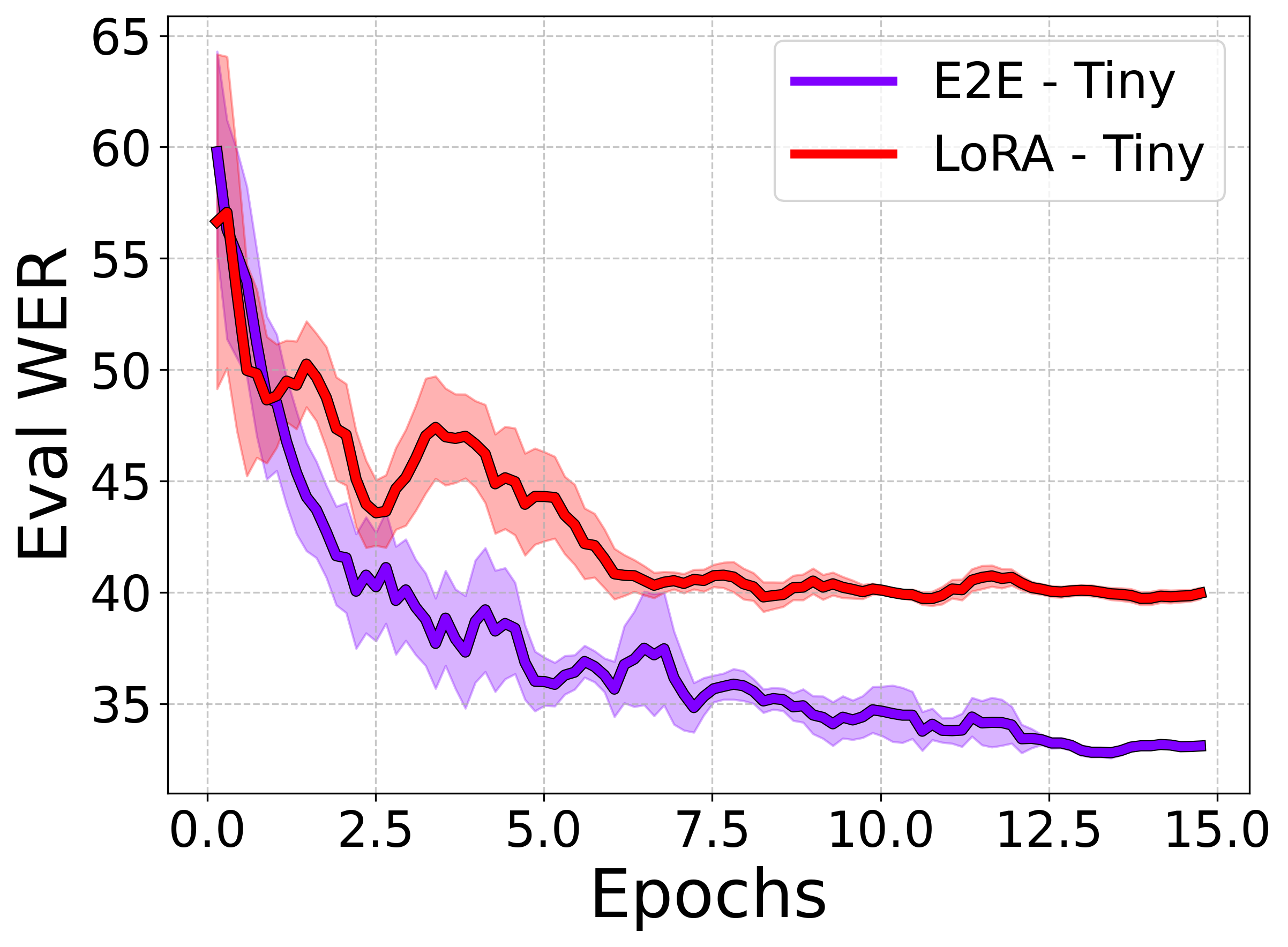}
        \caption{Tiny Train WER}
    \end{subfigure}
    \begin{subfigure}[b]{0.24\textwidth}
        \centering
        \includegraphics[width=\textwidth]{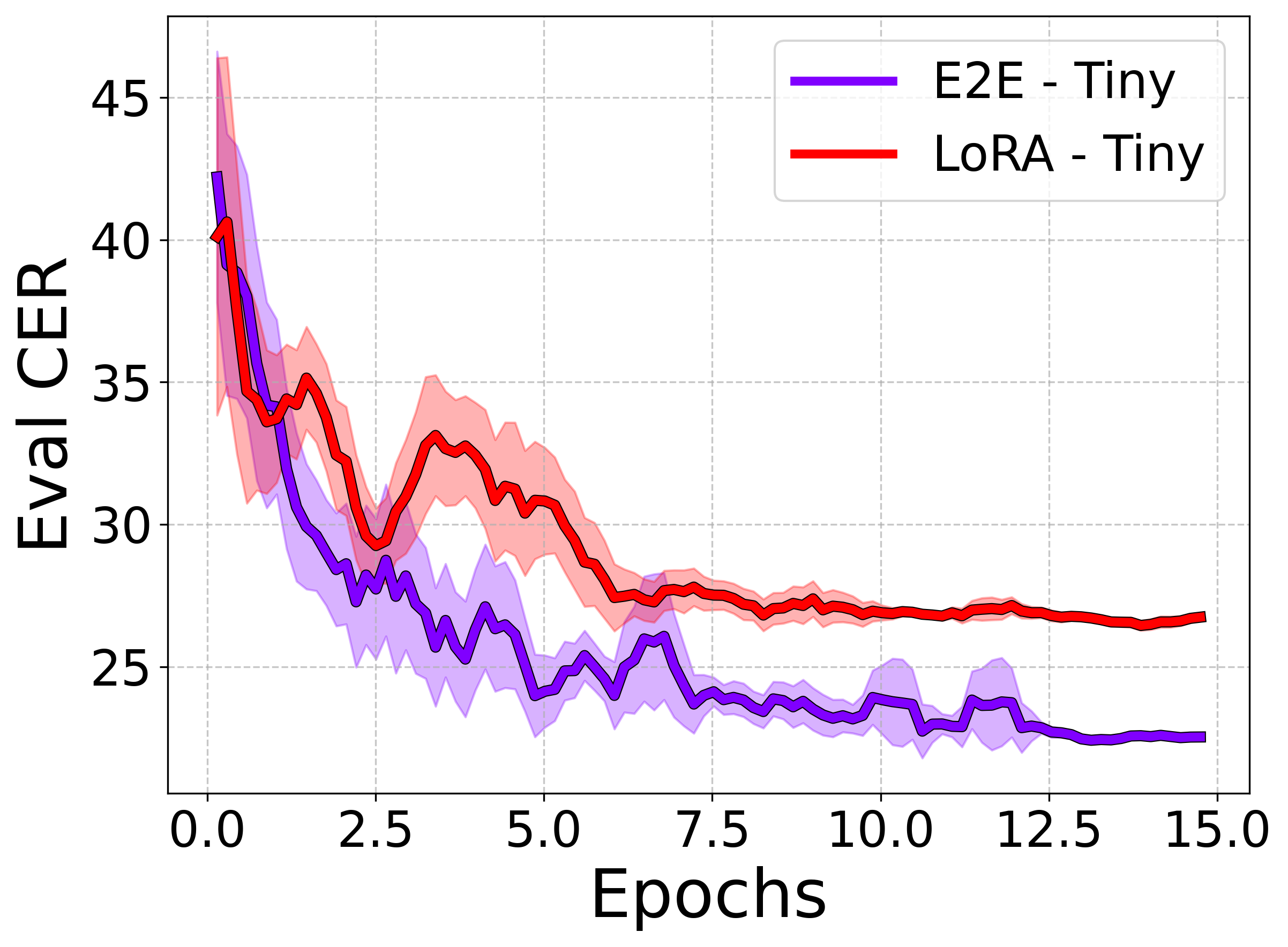}
        \caption{Tiny Eval CER}
    \end{subfigure}
    
    \begin{subfigure}[b]{0.24\textwidth}
        \centering
        \includegraphics[width=\textwidth]{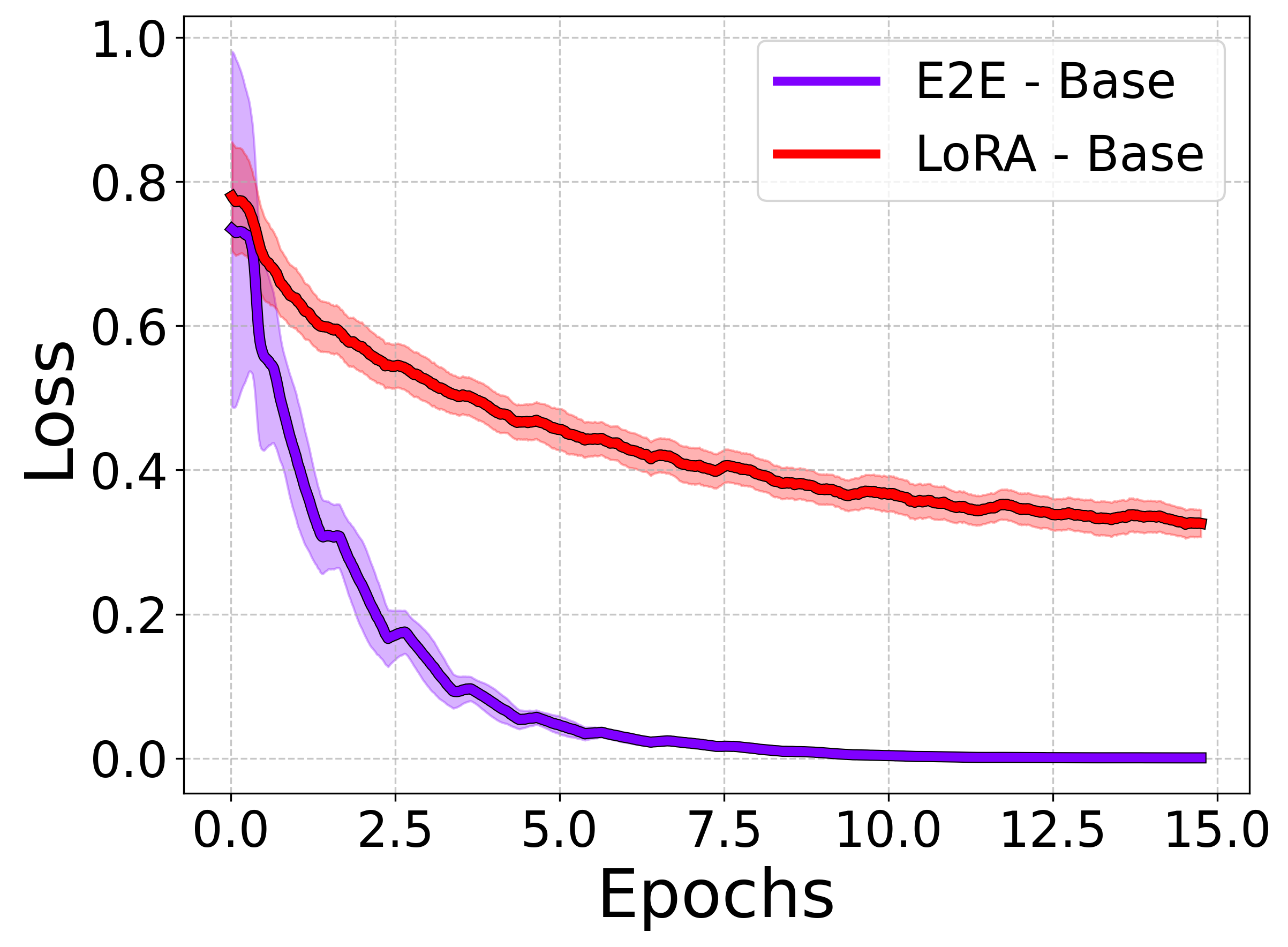}
        \caption{Base Train Loss}
    \end{subfigure}
    \begin{subfigure}[b]{0.24\textwidth}
        \centering
        \includegraphics[width=\textwidth]{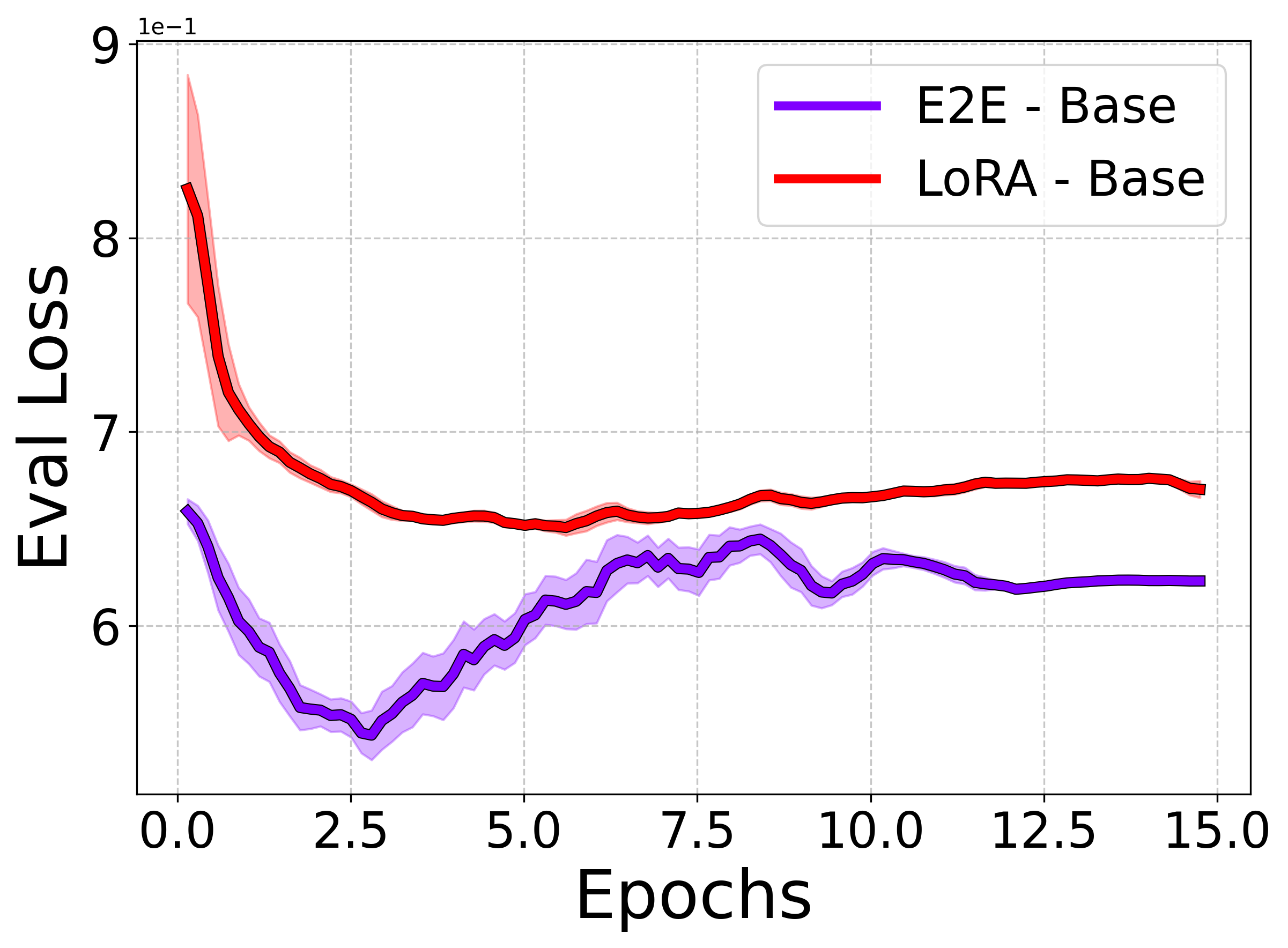}
        \caption{Base Eval Loss}
    \end{subfigure}
    \begin{subfigure}[b]{0.24\textwidth}
        \centering
        \includegraphics[width=\textwidth]{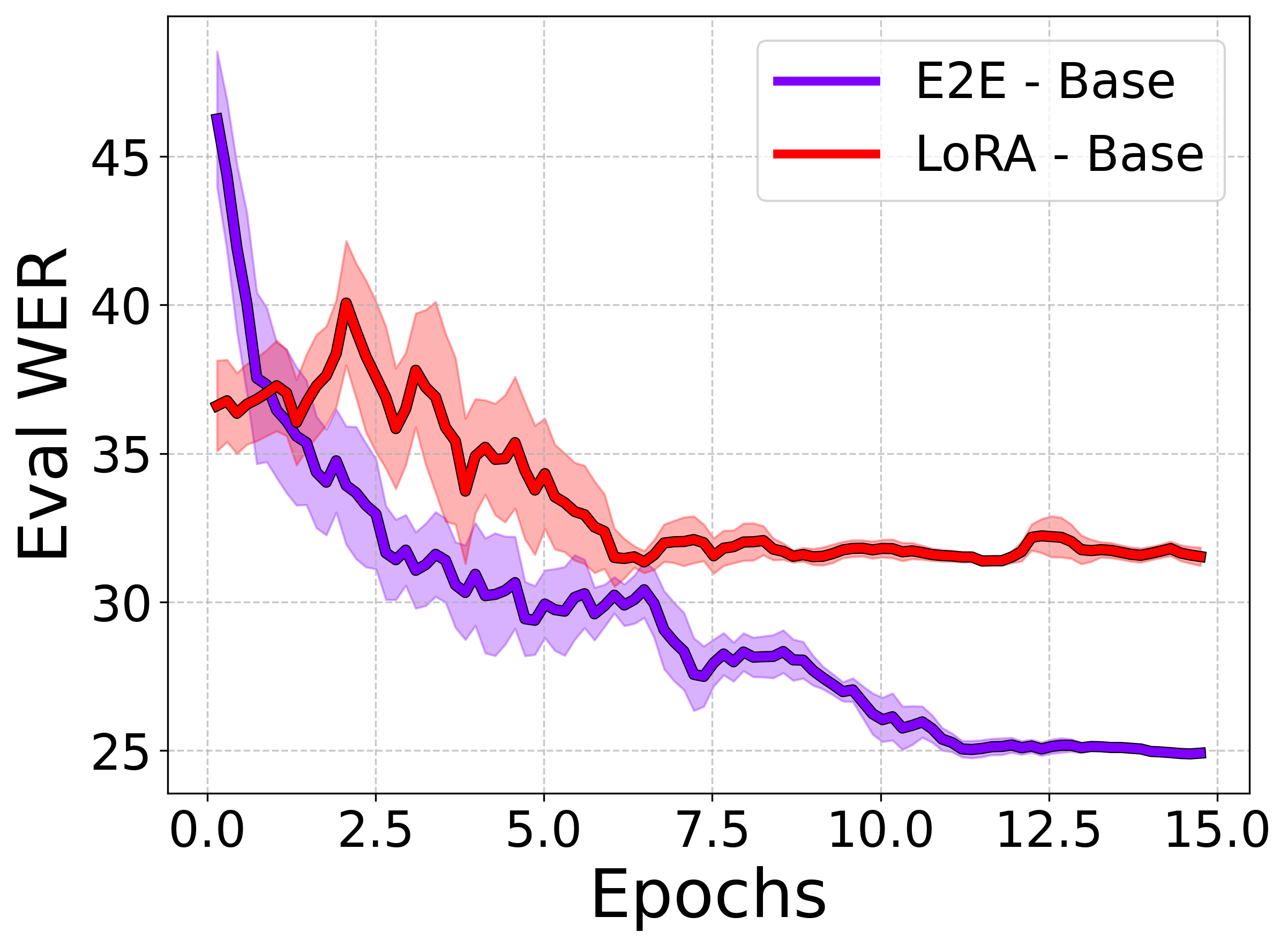}
        \caption{Base Train WER}
    \end{subfigure}
    \begin{subfigure}[b]{0.24\textwidth}
        \centering
        \includegraphics[width=\textwidth]{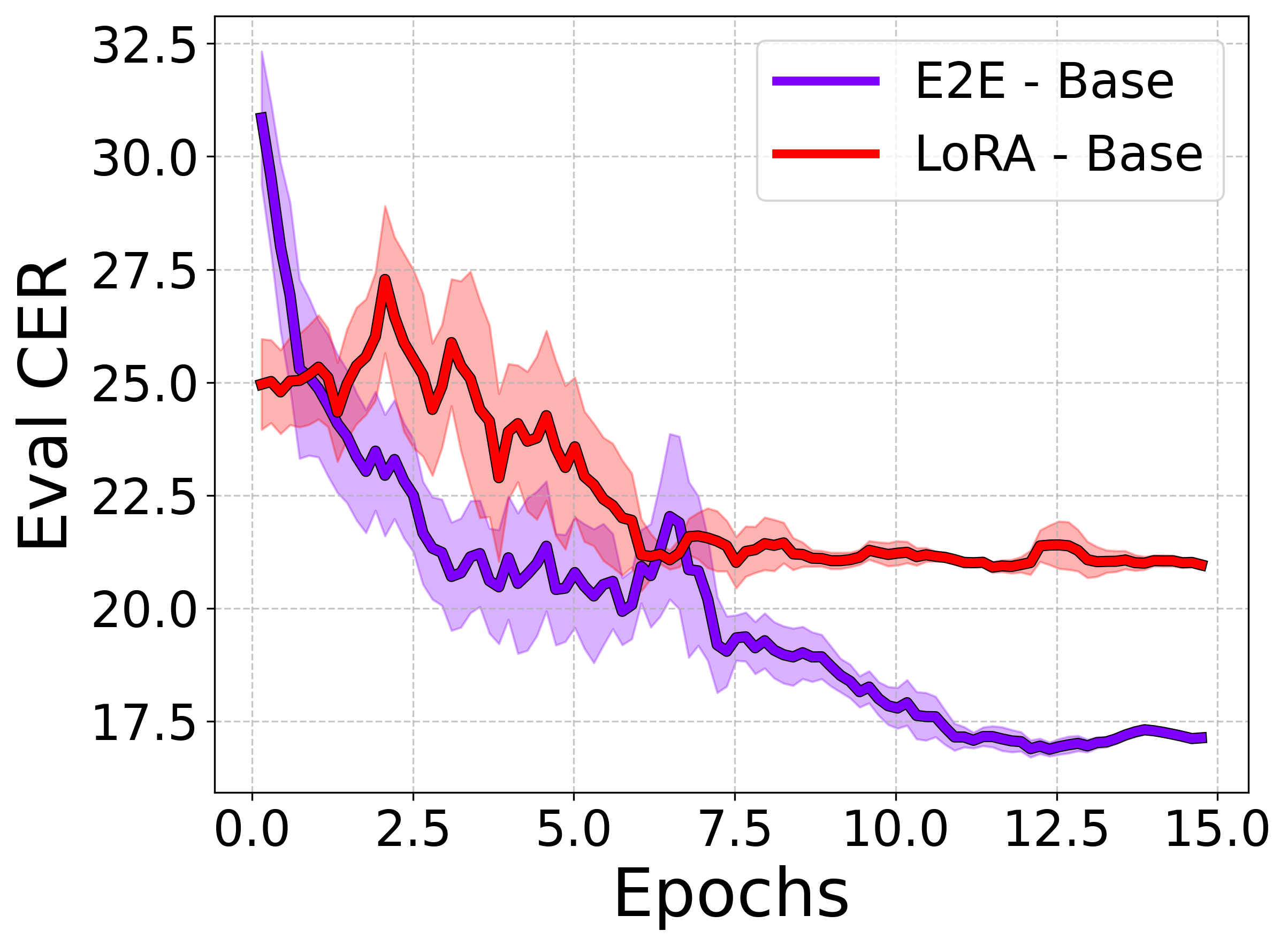}
        \caption{Base Eval CER}
    \end{subfigure}
    
    \begin{subfigure}[b]{0.24\textwidth}
        \centering
        \includegraphics[width=\textwidth]{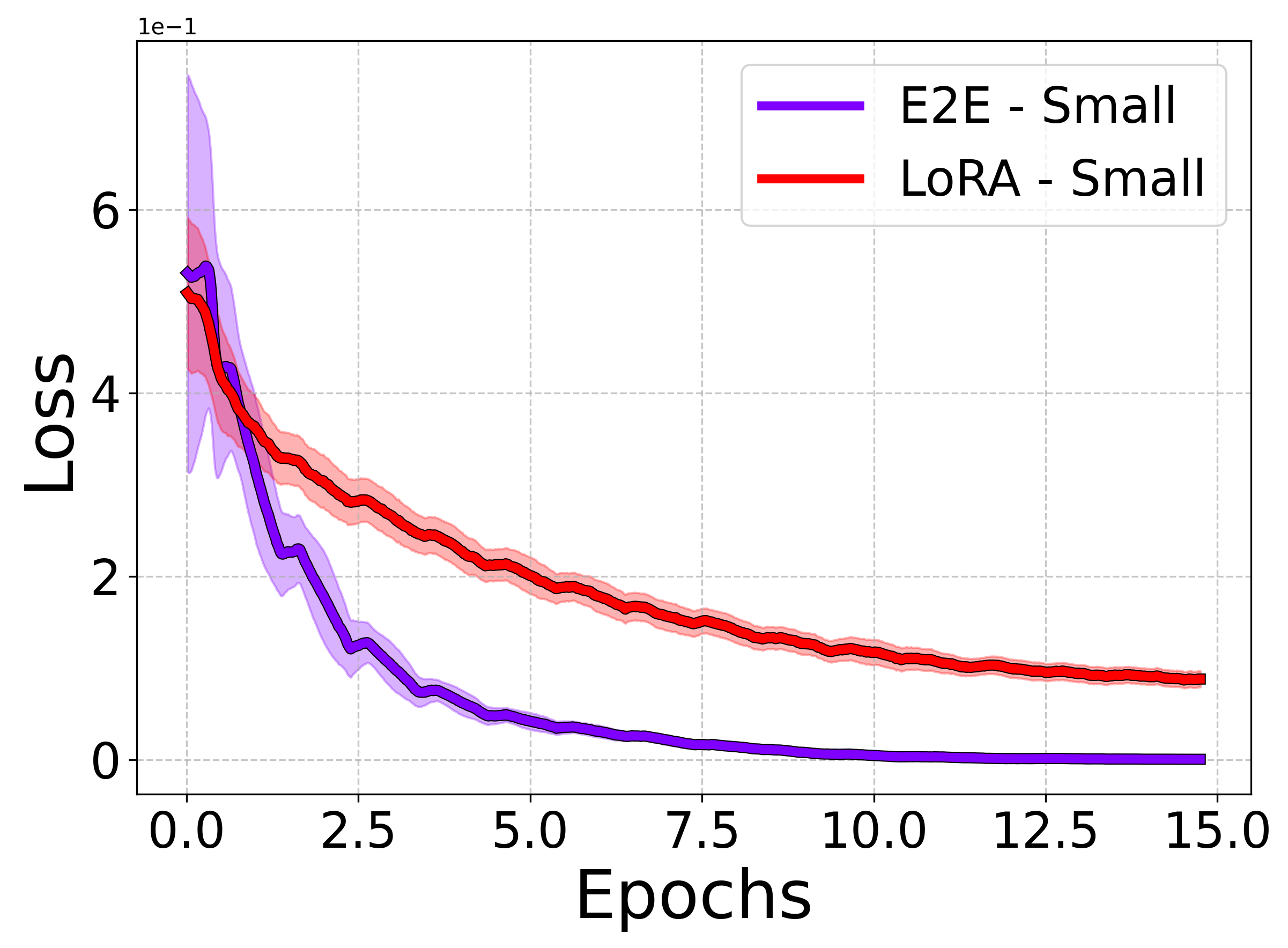}
        \caption{Small Train Loss}
    \end{subfigure}
    \begin{subfigure}[b]{0.24\textwidth}
        \centering
        \includegraphics[width=\textwidth]{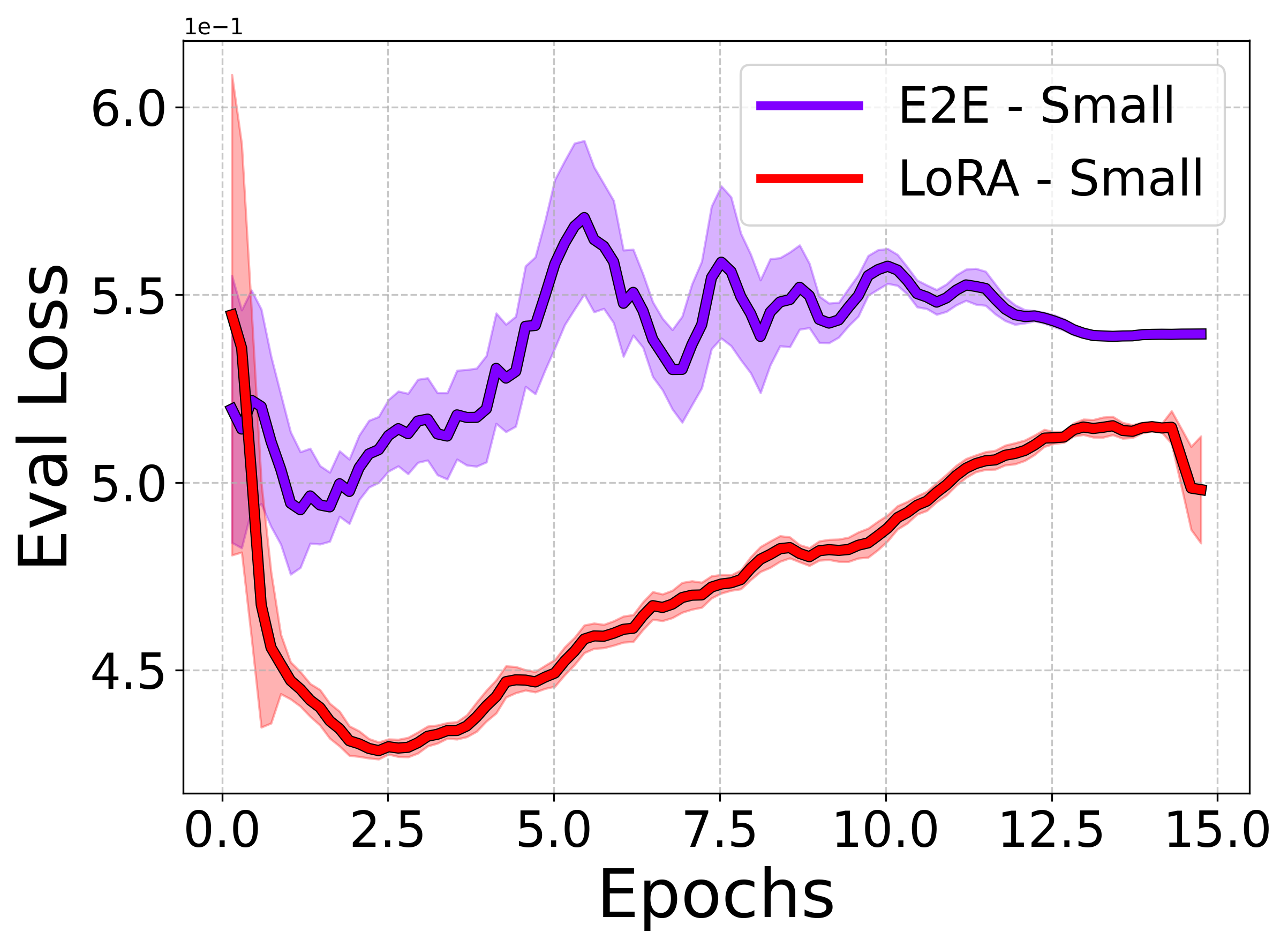}
        \caption{Small Eval Loss}
    \end{subfigure}
    \begin{subfigure}[b]{0.24\textwidth}
        \centering
        \includegraphics[width=\textwidth]{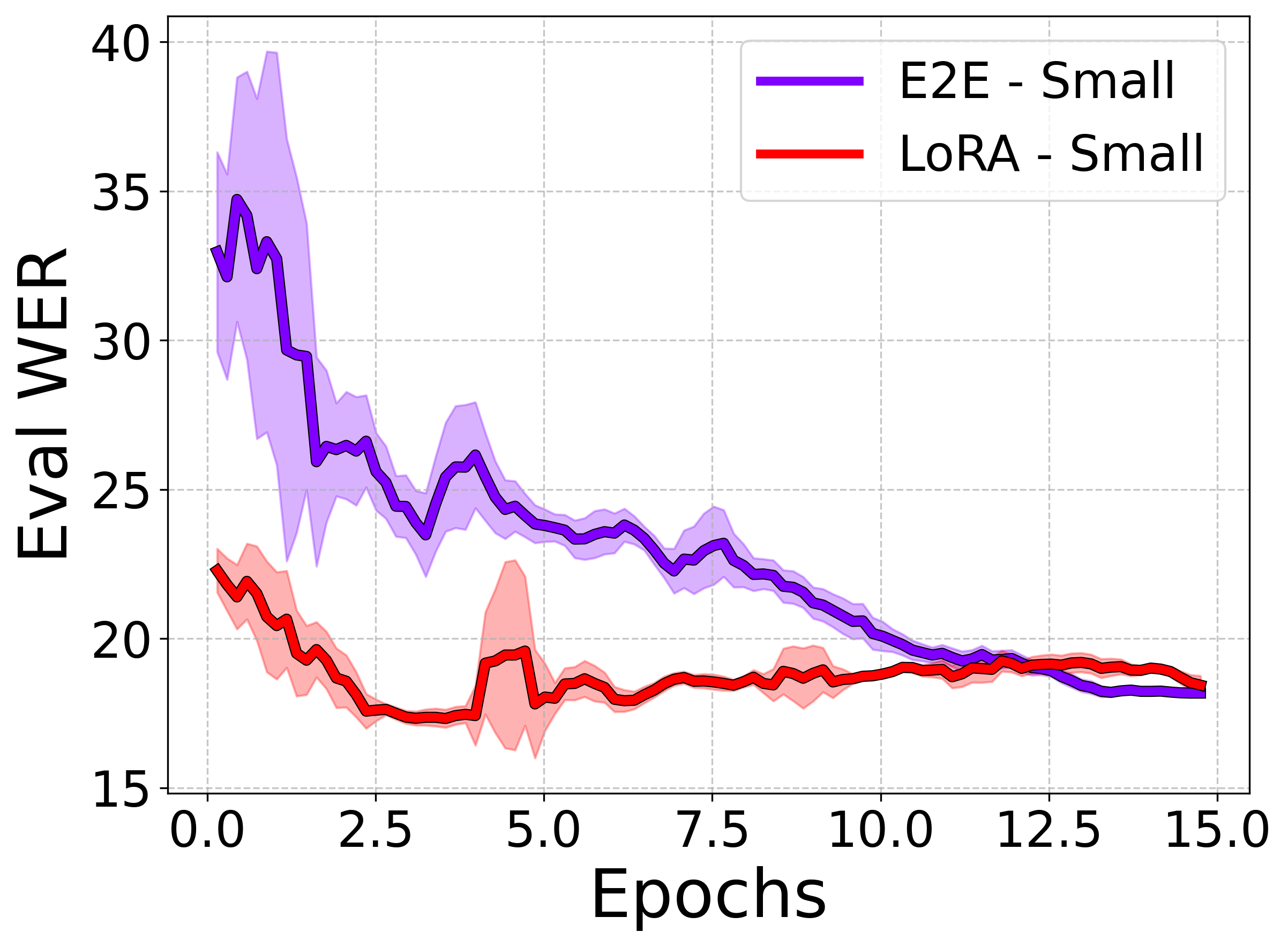}
        \caption{Small Train WER}
    \end{subfigure}
    \begin{subfigure}[b]{0.24\textwidth}
        \centering
        \includegraphics[width=\textwidth]{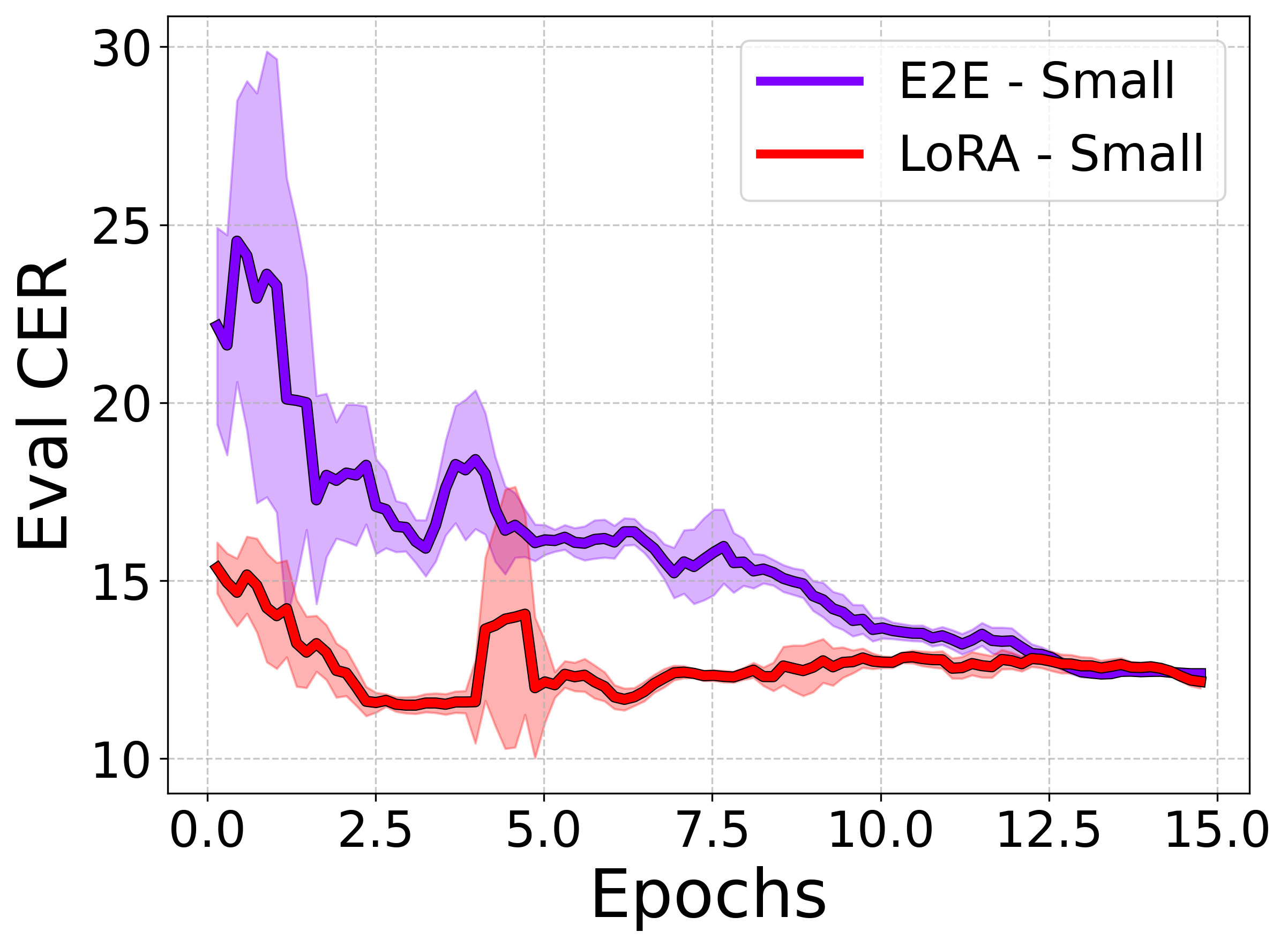}
        \caption{Small Eval CER}
    \end{subfigure}
    \caption{Training and evaluation metrics for Tiny (top row), Base (middle row), and Small (bottom row) models. The columns represent, from left to right: training loss, evaluation loss, training WER, and evaluation CER. For the Tiny and Base models, experiments demonstrate that E2E fine-tuning achieves superior performance. In contrast, the Small model exhibits better convergence with LoRA fine-tuning likely due to its ability to efficiently adapt large parameter sets while maintaining computational efficiency. Refer to subfigures j, k, and l for details.}
    \label{fig:training_metrics}
\end{figure*}

\textbf{Evaluation Metrics:}
We evaluate our Japanese ASR models using Word Error Rate (WER) and Character Error Rate (CER). WER assesses word-level accuracy, while CER captures character-level differences. Together, these metrics provide a comprehensive view of the model's performance.

WER and CER are calculated as follows:

\begin{equation}
\text{WER} = \frac{S + D + I}{N}
\end{equation}
\quad
\begin{equation}
\text{CER} = \frac{S + D + I}{C}
\end{equation}

\vspace{0.2cm}

\noindent where \(S\) is the number of substitutions (incorrect words or characters), \(D\) is the number of deletions (missing words or characters), and \(I\) is the number of insertions (extra words or characters). \(N\) represents the total number of words in the reference (used for Word Error Rate, WER), while \(C\) is the total number of characters in the reference (used for Character Error Rate, CER).



In Japanese, punctuation and spacing often lack semantic value but are treated as errors by WER and CER. To address this, we apply normalization to remove punctuation, unify spacing, and standardize full-width and half-width characters. These preprocessing steps ensure the metrics focus on meaningful differences rather than formatting inconsistencies.

However, upon examining our outputs, we noticed there were instances where words were spelled out in hiragana (the phonetic writing system) in the ground truth labels, while the model outputted the same word in kanji (the conceptual writing system based on Chinese characters), or vice versa. Such differences were counted as errors in both CER and WER metrics, but are not true errors because they are often interchangeable while maintaining semantic and phonetic correctness. This highlights a limitation of standard metrics generally applied across languages, as revealed by the complexities of the Japanese language, which remains an active area of research \cite{karitaLenientEvaluationJapanese2023}. We envision that other languages will also have idiosyncrasies that require nuanced evaluation.

\begin{table}[t]
    \centering
    \begin{tabular}{lcc}
    \toprule
    \textbf{Model} & \textbf{WER (\%)} & \textbf{CER (\%)} \\
    \midrule
    Whisper Tiny & 47.48 & 32.74 \\
    Whisper Base & 29.81 & 20.20 \\
    Whisper Small & 16.14 & 9.89 \\
    Whisper Medium & 10.84 & 6.86 \\
    Whisper Large & 7.06 & 4.63 \\
    \midrule
    Whisper Tiny + LoRA & 33.16 & 20.83 \\
    Whisper Base + LoRA & 23.36 & 14.50 \\
    Whisper Small + LoRA & 14.90 & 9.16 \\
    \midrule
    Whisper Tiny + End-to-End & 23.67 & 14.72 \\
    Whisper Base + End-to-End & 16.39 & 10.07 \\
    Whisper Small + End-to-End & 12.19 & 7.38 \\
    \midrule
    ReazonSpeech (NeMo) & 7.75 & 5.24 \\
    ReazonSpeech (K2) & 8.55 & 6.07 \\
    ReazonSpeech (ESPnet) & \textbf{7.05} & \textbf{4.62} \\
    \bottomrule
    \end{tabular}
    \caption{Comparison of WER and CER across different ASR models and fine-tuning approaches on Japanese speech recognition.}
    \label{tab:asr_results}
\end{table}

\textbf{Data Augmentation:}
We applied two types of data augmentation to training data: audio data augmentation and SpecAugment.
For audio augmentation, we experimented with techniques such as time-stretching, pitch alteration, gain adjustment, and adding Gaussian noise. However, the settings that we used were minimal and did not significantly affect the dataset. Combined with the fact that we already had a large amount of training data from four diverse datasets, audio augmentation did not improve performance and only increased training time by doubling the dataset size. Consequently, we decided to turn off audio augmentation and focus on SpecAugment instead.

\begin{figure}
    \centering
    \includegraphics[width=1.0\linewidth]{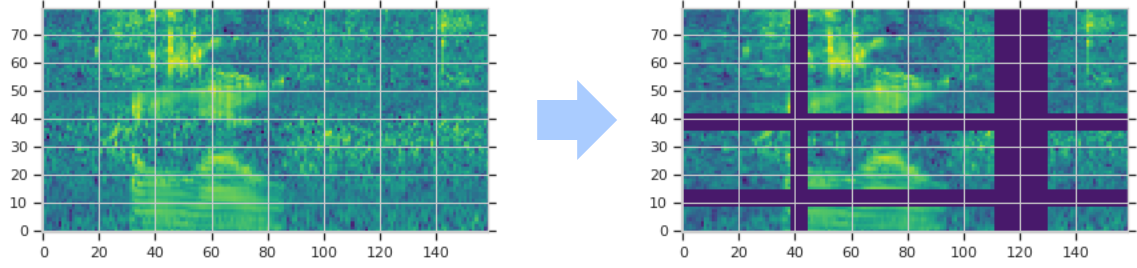}
    \caption{Visualization of SpecAugment: Original Log-Mel spectrum (Left) vs Augmented Spectrogram with Time and Frequency Masking (Right)}
    \label{fig:spec-augment}
\end{figure}

\begin{figure}[t]
    \centering
    \includegraphics[width=0.8\linewidth]{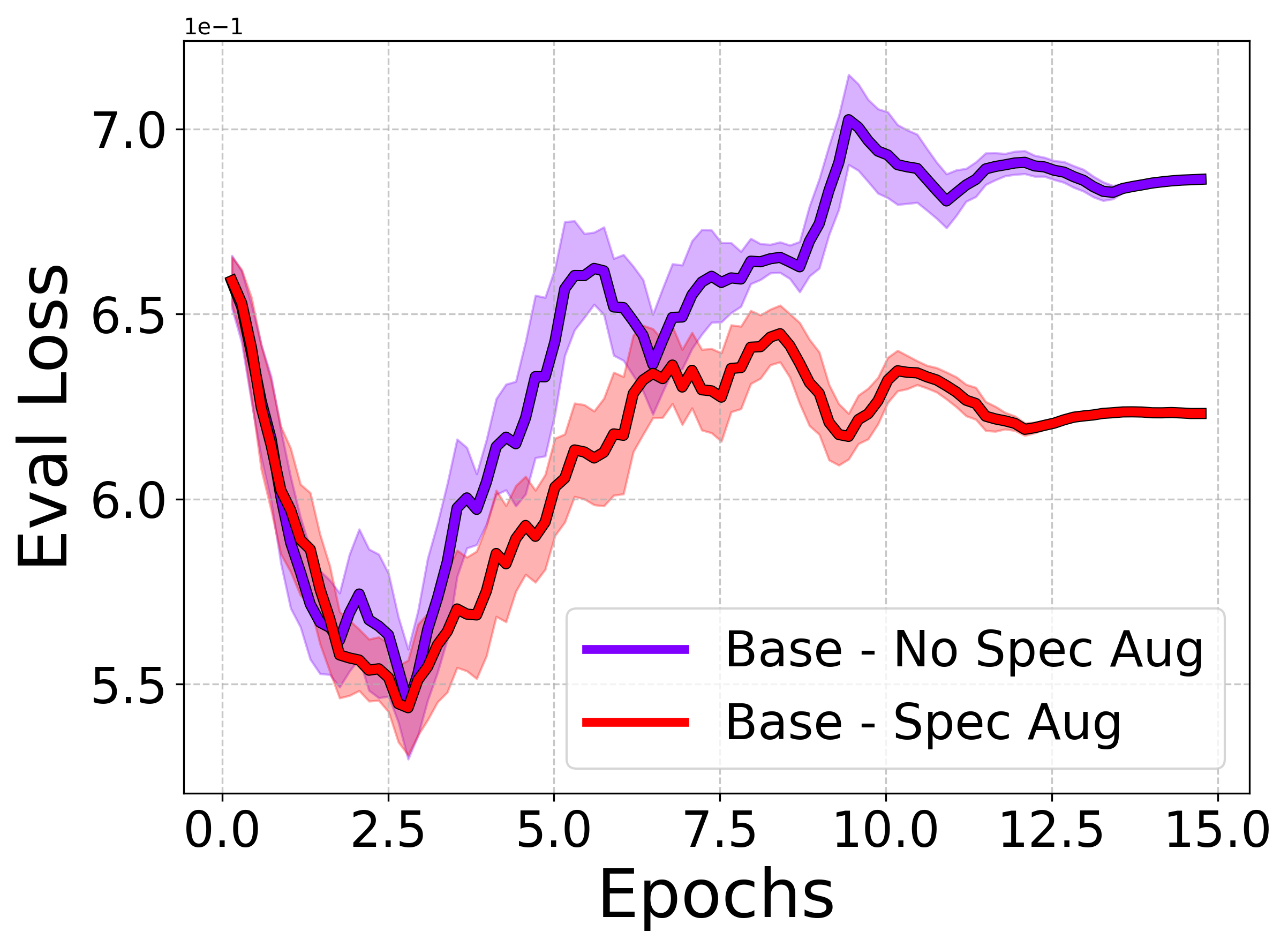}
    \caption{Evaluation loss comparison between models trained with and without SpecAugment}
    \label{fig:spec_aug_comparison}
\end{figure}

SpecAugment, which involves modifying the spectrograms by randomly masking frequency bands and time intervals to improve robustness(Figure~\ref{fig:spec-augment})\cite{SpecAugment}, showed clear benefits. The results demonstrate that SpecAugment significantly improves model performance by reducing overfitting and enhancing generalization capabilities(Figure~\ref{fig:spec_aug_comparison}). This highlights the effectiveness of SpecAugment in enhancing model performance.

\begin{figure*}[ht]
    \centering
    \includegraphics[width=0.95\linewidth]{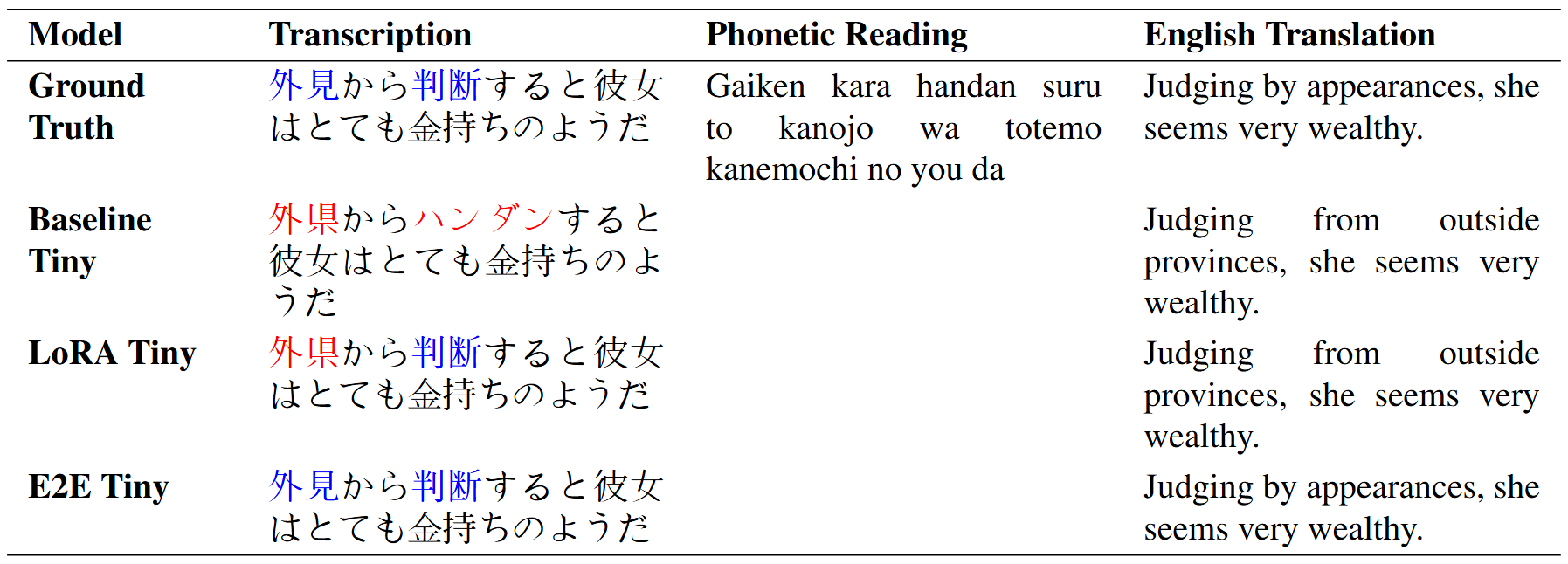}
    \caption{Transcription samples from various models with their phonetic readings and English translations. The phonetic readings are consistent across all models, but differences in Kanji lead to significant variations in semantic meaning.}
    \label{tab:transcription_comparison}
\end{figure*}

\textbf{Finding an Optimal LoRA Rank:}
In Figure~\ref{fig:lora_comparison}, we compare the performance of LoRA fine-tuning across different ranks. The rank corresponds to the number of trainable parameters in the LoRA adapter layers, which are combined with the original model parameters to produce the final output during inference.

Our experiments considered ranks ranging from 64 to 256. We observe that increasing the rank leads to improved model performance, approaching the results of full end-to-end fine-tuning. This trend suggests that higher ranks enable the LoRA adapters to better capture the complexity inherent in the dataset.

Among the tested configurations, a rank of 256 yielded the best performance. This is likely due to the dataset's large size and diversity, necessitating a higher rank to capture its complexity effectively. While higher ranks could potentially improve performance further, constraints on GPU memory and training time limited our experiments to a maximum rank of 256.

\subsection{Quantitative Analysis}
The results demonstrate significant improvements in CER as models undergo fine-tuning and increase in size. The Original Tiny model, starting with the highest CER at 32.7, highlights the limitations of the base Tiny architecture in its unoptimized state. However, the introduction of LoRA fine-tuning dramatically reduces the CER to 20.8, bringing its performance on par with the Original Base model (CER: 20.2). This illustrates the effectiveness of LoRA in narrowing the performance gap between smaller and larger models through targeted optimization.

Further enhancement is seen in the E2E Tiny model, which achieves a CER of 14.7, outperforming the Original Base model. This result underscores the capability of end-to-end fine-tuning to leverage the Tiny model’s architecture more effectively, surpassing the baseline of a larger model.

Finally, the Original Small model achieves the lowest CER of 9.9, reflecting the inherent advantages of increased model size in handling transcription tasks. However, the progression observed in the Tiny models demonstrates that with strategic fine-tuning, even smaller architectures can achieve results competitive with, and in some cases superior to, larger models. This finding emphasizes the value of fine-tuning techniques like LoRA and E2E in maximizing the potential of resource-constrained ASR systems.

\subsection{Qualitative Analysis}
This section provides a qualitative exploration of ASR model performance, focusing on how fine-tuning approaches, such as LoRA and E2E, introduce improvements over baseline models. We highlight two representative cases from our analysis: the first showcases successful adaptation to contextual nuances enabled by fine-tuning, while the second illustrates limitations in handling domain-specific terminology when such terms are not adequately represented in the fine-tuning data.

\textbf{Handling Ambiguity in Contextual Judgments:}
Here, we evaluate how ASR models address ambiguity in transcription when faced with phonetically similar phrases. Accurate recognition in such cases relies heavily on contextual clues. The phonetic readings produced by all models (Tiny Base, Tiny LoRA, and Tiny E2E) are identical, reflecting that the models consistently capture the correct pronunciation of the input. However, Tiny LoRA and Tiny E2E exhibit a notable advancement over Tiny Base by generating kanji characters that more accurately reflect the intended meaning within the context of the sentence. This improvement highlights the effectiveness of fine-tuning techniques, which enable the models to go beyond merely preserving phonetic fidelity and instead incorporate semantic understanding to produce contextually appropriate transcriptions. These results emphasize how targeted adjustments can refine a model's ability to align written output with both phonetic and contextual correctness(see Figure~\ref{tab:transcription_comparison} for details).

\textbf{Handling of Technical Terms and Specialized Vocabulary:}
An area where the models encountered notable challenges was in handling specialized medical terms. Terms such as "encephalitis" and "Listeria" were frequently misinterpreted across all models. Although fine-tuning provided minor improvements, these errors surface the limitations of the current training data and the necessity of incorporating domain-specific datasets to improve recognition. Additional details in the Appendix (Figure~\ref{tab:transcription_comparison_appendix}).

These findings underscore the varying capabilities of ASR models in handling ambiguity and specialized vocabulary, emphasizing the need for targeted improvements in context understanding and domain-specific training. 

\section{Conclusion}
This study explores the potential of language-specific fine-tuning of general multi-lingual ASR models like Whisper, to benefit from the learned base of language-agnostic patterns before specializing in a single language. 
We showed that by utilizing Japanese-specific datasets and employing LoRA for parameter-efficient updates or end-to-end fine-tuning we were able to improve Whisper's ASR accuracy while maintaining its resource-efficient architecture. 
The significance of the improvement in ASR quality from fine-tuning cannot be understated, as we were able to show that a fine-tuned Whisper Tiny model could achieve higher performance than the larger baseline Whisper-Base model. This research contributes a scalable and resource-efficient approach for improving ASR systems that leverage publicly available multi-lingual models in resource-constrained settings.
Fine-tuning multi-lingual models to achieve language-specific specialization offers an inclusive pathway for diverse linguistic communities, especially in languages that may lack the volume of data needed to train effectively train a mono-lingual model.

\clearpage

{\small
\bibliographystyle{unsrt}
\bibliography{egbib}

\begin{thebibliography}{10}

\bibitem{huggingfaceReazonspeechDatasets}
Reazon Human~Interaction Lab.
\newblock Reazonspeech - datasets at hugging face.

\bibitem{radford2023robust}
Alec Radford, Jong~Wook Kim, Tao Xu, Greg Brockman, Christine McLeavey, and Ilya Sutskever.
\newblock Robust speech recognition via large-scale weak supervision.
\newblock In {\em International conference on machine learning}, pages 28492--28518. PMLR, 2023.

\bibitem{githubWhispermodelcardmdMain}
openai.
\newblock whisper/model-card.md, 2024.

\bibitem{huLoRALowRankAdaptation2021}
Edward~J. Hu, Yelong Shen, Phillip Wallis, Zeyuan Allen-Zhu, Yuanzhi Li, Shean Wang, Lu~Wang, and Weizhu Chen.
\newblock {{LoRA}}: {{Low-Rank Adaptation}} of {{Large Language Models}}, 2021.

\bibitem{githubReazonSpeech}
Massive open japanese speech corpus.
\newblock [Accessed 09-12-2024].

\bibitem{githubOpenaiWhisper}
Github - openai/whisper.

\bibitem{conneauFLEURSFewshotLearning2022}
Alexis Conneau, Min Ma, Simran Khanuja, Yu~Zhang, Vera Axelrod, Siddharth Dalmia, Jason Riesa, Clara Rivera, and Ankur Bapna.
\newblock {{FLEURS}}: {{Few-shot Learning Evaluation}} of {{Universal Representations}} of {{Speech}}, 2022.

\bibitem{ardilaCommonVoiceMassivelyMultilingual2020}
Rosana Ardila, Megan Branson, Kelly Davis, Michael Henretty, Michael Kohler, Josh Meyer, Reuben Morais, Lindsay Saunders, Francis~M. Tyers, and Gregor Weber.
\newblock Common {{Voice}}: {{A Massively-Multilingual Speech Corpus}}, 2020.

\bibitem{sonobeJSUTCorpusFree2017}
Ryosuke Sonobe, Shinnosuke Takamichi, and Hiroshi Saruwatari.
\newblock {{JSUT}} corpus: Free large-scale {{Japanese}} speech corpus for end-to-end speech synthesis, 2017.

\bibitem{yinReazonSpeechFreeMassive}
Yue Yin, Daijiro Mori, and Seiji Fujimoto.
\newblock {{ReazonSpeech}}: {{A Free}} and {{Massive Corpus}} for {{Japanese ASR}}, 2023.

\bibitem{radfordRobustSpeechRecognition}
Alec Radford, Jong~Wook Kim, Tao Xu, Greg Brockman, Christine McLeavey, and Ilya Sutskever.
\newblock Robust speech recognition via large-scale weak supervision, 2022.

\bibitem{holdingsReazonSpeechV21Setting2024}
Yue Yin, Daijiro Mori, and Seiji Fujimoto.
\newblock Reazonspeech v2.1: Setting a new standard in japanese asr, 2024.

\bibitem{IndroducingWhisper}
OpenAI.
\newblock Introducing whisper.
\newblock GitHub Repository, 2022.

\bibitem{MediumWhisper}
David Cochard.
\newblock Whisper: Speech recognition model capable of recognizing 99 languages.
\newblock Medium, 2023.

\bibitem{ReazonSpeechNemo}
Reazon Human~Interaction Lab.
\newblock Reazonspeechnemo hugging face repository.
\newblock Hugging Face Repository.

\bibitem{ReazonSpeechK2}
Reazon Human~Interaction Lab.
\newblock Reazonspeechk2 hugging face repository.
\newblock Hugging Face Repository.

\bibitem{ReazonSpeechESPNet}
Reazon Human~Interaction Lab.
\newblock Reazonspeechespnet hugging face repository.
\newblock Hugging Face Repository.

\bibitem{hu2021loralowrankadaptationlarge}
Edward~J. Hu, Yelong Shen, Phillip Wallis, Zeyuan Allen-Zhu, Yuanzhi Li, Shean Wang, Lu~Wang, and Weizhu Chen.
\newblock Lora: Low-rank adaptation of large language models, 2021.

\bibitem{xu2023parameterefficientfinetuningmethodspretrained}
Lingling Xu, Haoran Xie, Si-Zhao~Joe Qin, Xiaohui Tao, and Fu~Lee Wang.
\newblock Parameter-efficient fine-tuning methods for pretrained language models: A critical review and assessment, 2023.

\bibitem{karitaLenientEvaluationJapanese2023}
Shigeki Karita, Richard Sproat, and Haruko Ishikawa.
\newblock Lenient {{Evaluation}} of {{Japanese Speech Recognition}}: {{Modeling Naturally Occurring Spelling Inconsistency}}.
\newblock In {\em Proceedings of the {{Workshop}} on {{Computation}} and {{Written Language}} ({{CAWL}} 2023)}, pages 61--70. Association for Computational Linguistics, 2023.

\bibitem{SpecAugment}
Daniel~S Park, William Chan, Yu~Zhang, Chung-Cheng Chiu, Barret Zoph, Ekin~D Cubuk, and Quoc~V Le.
\newblock Specaugment: A simple data augmentation method for automatic speech recognition.
\newblock {\em arXiv preprint arXiv:1904.08779}, 2019.

\end{thebibliography}
}

\clearpage

\appendix
\renewcommand{\thesection}{\Alph{section}}
\setcounter{section}{0}

\onecolumn

\section*{Appendix}

\section{Handling of Technical Terms and Specialized Vocabulary}

\begin{figure}[htbp]
    \centering
    \includegraphics[width=0.9\linewidth]{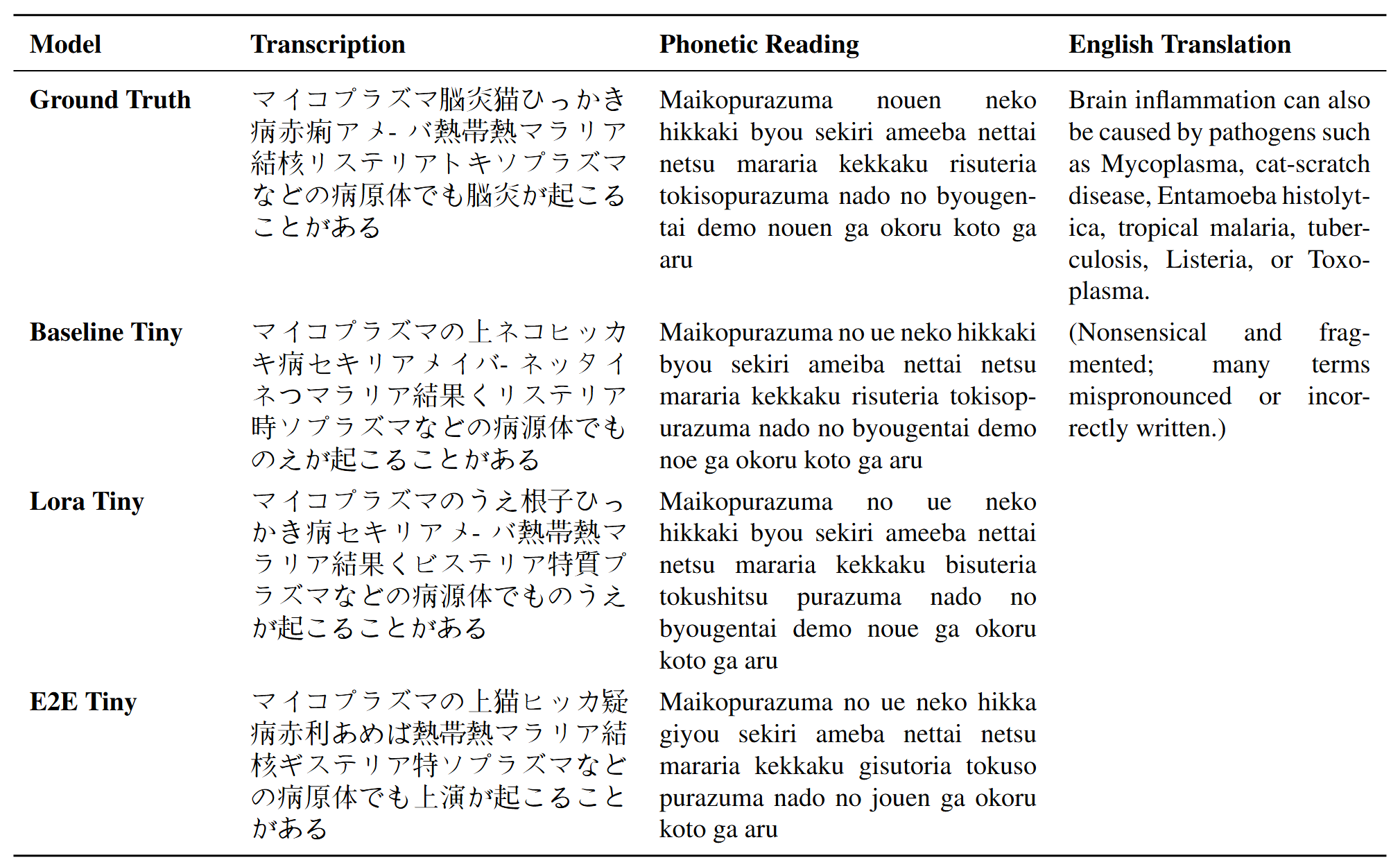}
    \caption{Additional qualitative samples from our trained model.}
    \label{tab:transcription_comparison_appendix}
\end{figure}

\section{Source Code}

\noindent Our source code is available at:  
\url{https://github.com/ryujimorita/tokyo_whisperers}

\twocolumn

\end{document}